\RequirePackage[svgnames]{xcolor}

\documentclass[11pt,letterpaper]{mystyle}

\usepackage[all]{hypcap}
\usepackage[svgnames]{xcolor}
\usepackage[comma,authoryear,compress]{natbib}
\usepackage{hyperref}[citecolor=lightblue]

\hypersetup{
    colorlinks = true,
    citecolor = {YaleBlue},
}

\usepackage{algorithm}
\usepackage{algorithmicx}
\usepackage{algpseudocode}
\usepackage{microtype}
\usepackage{graphicx}
\expandafter\def\csname ver@subfig.sty\endcsname{}
\usepackage{booktabs} %
\usepackage{float}
\usepackage{bigstrut}

\usepackage{amsmath}
\usepackage{amssymb}
\usepackage{mathtools}
\usepackage{amsthm}
\usepackage{mathrsfs}
\usepackage{nicefrac}
\usepackage{dsfont}
\usepackage{enumitem}
\usepackage{subcaption}
\usepackage{graphicx,subfig}
\usepackage{cleveref}
\usepackage{bxcoloremoji}

\usepackage{siunitx}
\usepackage{booktabs}
\usepackage{tabularx}
\usepackage{colortbl}
\usepackage{makecell}
\usepackage{array}

\usepackage{float}

\usepackage[utf8]{inputenc} %
\usepackage[T1]{fontenc}    %
\usepackage{hyperref}       %
\usepackage{url}            %
\usepackage{booktabs}       %
\usepackage{amsfonts}       %
\usepackage{nicefrac}       %
\usepackage{microtype}      %
\usepackage{graphicx}
\usepackage{subcaption} 
\usepackage{amssymb}
\usepackage{fdsymbol}
\usepackage{wrapfig}
\usepackage{lipsum}
\usepackage{enumitem}
\usepackage{stackengine}
\usepackage[font=small,labelfont=bf]{caption}
\usepackage{color}
\usepackage{adjustbox}

\usepackage{rotating}
\usepackage{makecell}

\newcommand{\x}{\mathbf{x}}

\definecolor{blanchedalmond}{rgb}{1.0, 0.92, 0.8}
\definecolor{carmine}{rgb}{0.59, 0.0, 0.09}
\definecolor{lightblue}{rgb}{0.22,0.45,0.70}%

\renewcommand{\mathbf}{\boldsymbol}

\makeatletter
\def\Ddots{\mathinner{\mkern1mu\raise\p@
\vbox{\kern7\p@\hbox{.}}\mkern2mu
\raise4\p@\hbox{.}\mkern2mu\raise7\p@\hbox{.}\mkern1mu}}
\makeatother

\definecolor{amaranth}{rgb}{0.9, 0.17, 0.31}
\definecolor{antiquebrass}{rgb}{0.8, 0.58, 0.46}
\definecolor{antiquefuchsia}{rgb}{0.57, 0.36, 0.51}
\definecolor{chromeyellow}{rgb}{0.31, 0.47, 0.26}

\newtcolorbox{AIbox}[2][]{aibox,title=#2,#1}
\definecolor{lightblue}{rgb}{0.22,0.45,0.70}%
\definecolor{Gray}{gray}{0.95}
\definecolor{Cornsilk}{rgb}{1.0, 0.97, 0.86}

\usepackage{amsmath}

\usepackage[all]{hypcap}

\title{FlowVLA: Visual Chain of Thought-based Motion Reasoning for Vision-Language-Action Models}

\runningtitle{FlowVLA: Visual Chain of Thought-based Motion Reasoning for Vision-Language-Action Models}

\author{
  Zhide Zhong$^1$,
  Haodong Yan$^1$,
  Junfeng Li$^1$,
  Xiangchen Liu$^1$,
  Xin Gong$^1$,
  Tianran Zhang$^1$,
  Wenxuan Song$^1$,
  Jiayi Chen$^1$,
  Xinhu Zheng$^1$,
  Hesheng Wang$^2$, and
  Haoang Li
}

\affil[1]{HKUST(GZ)}
\affil[2]{Shanghai Jiao Tong University}

\begin{document}

\begin{abstract}
Many Vision-Language-Action (VLA) models are built upon an internal world model trained via next-frame prediction ``$v_t \rightarrow v_{t+1}$''. However, this paradigm attempts to predict the future frame's appearance directly, without explicitly reasoning about the underlying dynamics. This lack of an explicit motion reasoning step often leads to physically implausible visual forecasts and inefficient policy learning. To address this limitation, we introduce the \textbf{Visual Chain of Thought}, a paradigm that compels the model to first reason about motion dynamics before generating the future frame. We instantiate this paradigm by proposing \textbf{FlowVLA}, an autoregressive Transformer that explicitly materializes this reasoning process as ``$v_t \rightarrow f_t \rightarrow v_{t+1}$'', where $f_t$ is an intermediate optical flow prediction that inherently encodes motion. By forcing the model to first follow the motion plan encoded by $f_t$, this process aligns the pre-training objective of dynamics prediction with the downstream task of action generation. We conduct experiments on challenging robot manipulation benchmarks, as well as a real-robot platform. Our FlowVLA not only generates more coherent and physically plausible visual predictions, but also achieves state-of-the-art policy performance with substantially improved sample efficiency, pointing toward a more principled foundation for world modeling in VLAs. Project page: https://irpn-lab.github.io/FlowVLA/

\vspace{2mm}


\vspace{5mm}

\end{abstract}

\maketitle
\vspace{3mm}
\vspace{-4mm}
\section{Introduction}

\begin{figure*}[t!]
    \centering
    \includegraphics[width=\textwidth]{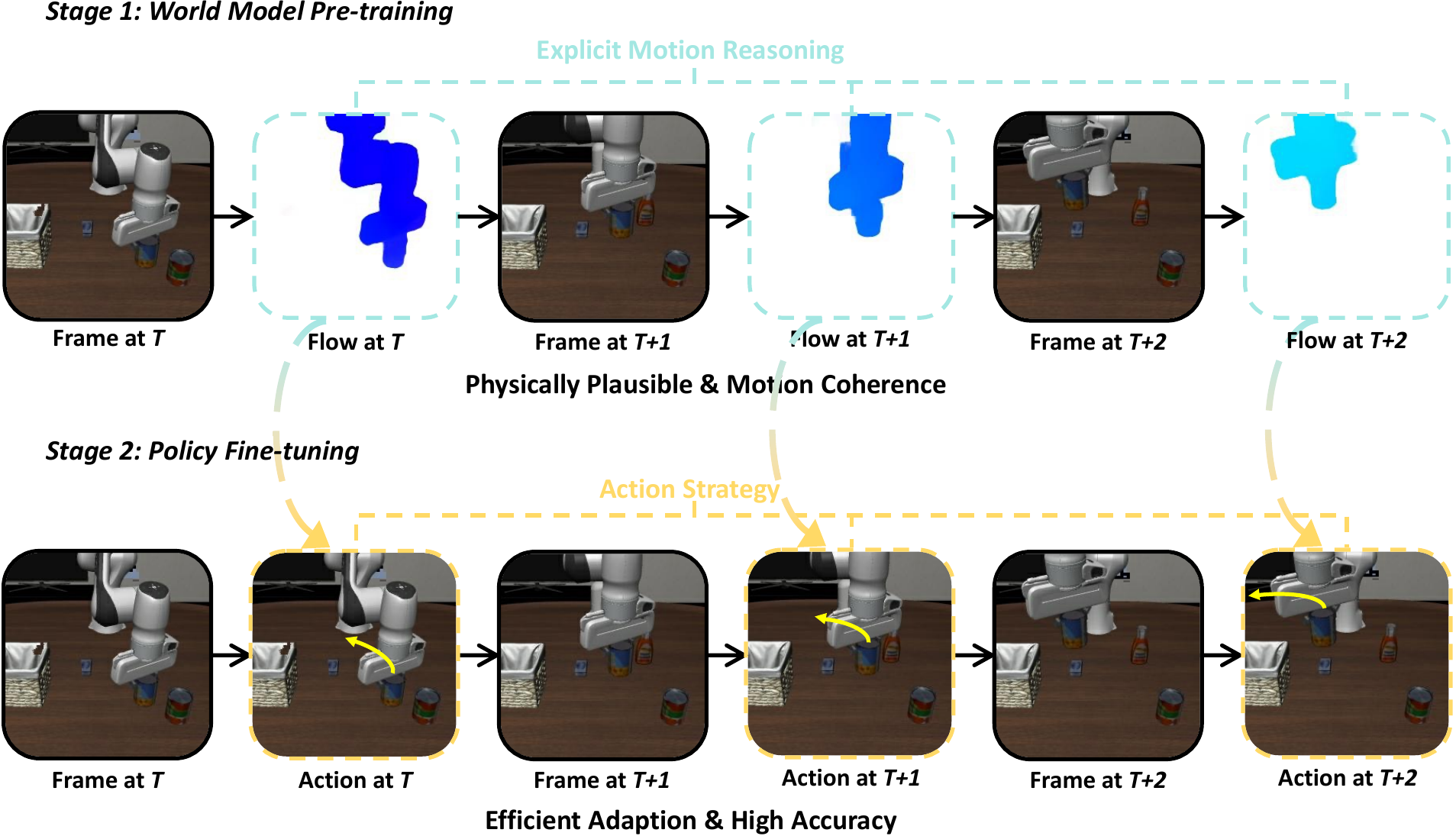}
    \caption{
    \textbf{Two-Stage Training Paradigm of FlowVLA.} 
    \textbf{(Top) Stage 1: World Model Pre-training with Visual CoT.} The model learns to predict an intermediate motion representation (\text{Flow}\ at\ $T$) from an initial frame (\text{Frame}\ at\ $T$), and then forecasts the subsequent frame (\text{Frame}\ at\ ${T+1}$). This iterative process yields physically plausible, long-horizon video predictions.
    \textbf{(Bottom) Stage 2: Policy Fine-tuning.} Through fine-tuning, the pre-trained world model is adapted to generate precise robot action chunk (\text{Action}\ at\ $T$) from visual observations. This paradigm leverages the learned dynamics for efficient and accurate policy learning.
    }
    \label{fig:teaser}
\end{figure*}

\begin{figure*}[t!]
    \centering
    \includegraphics[width=\textwidth]{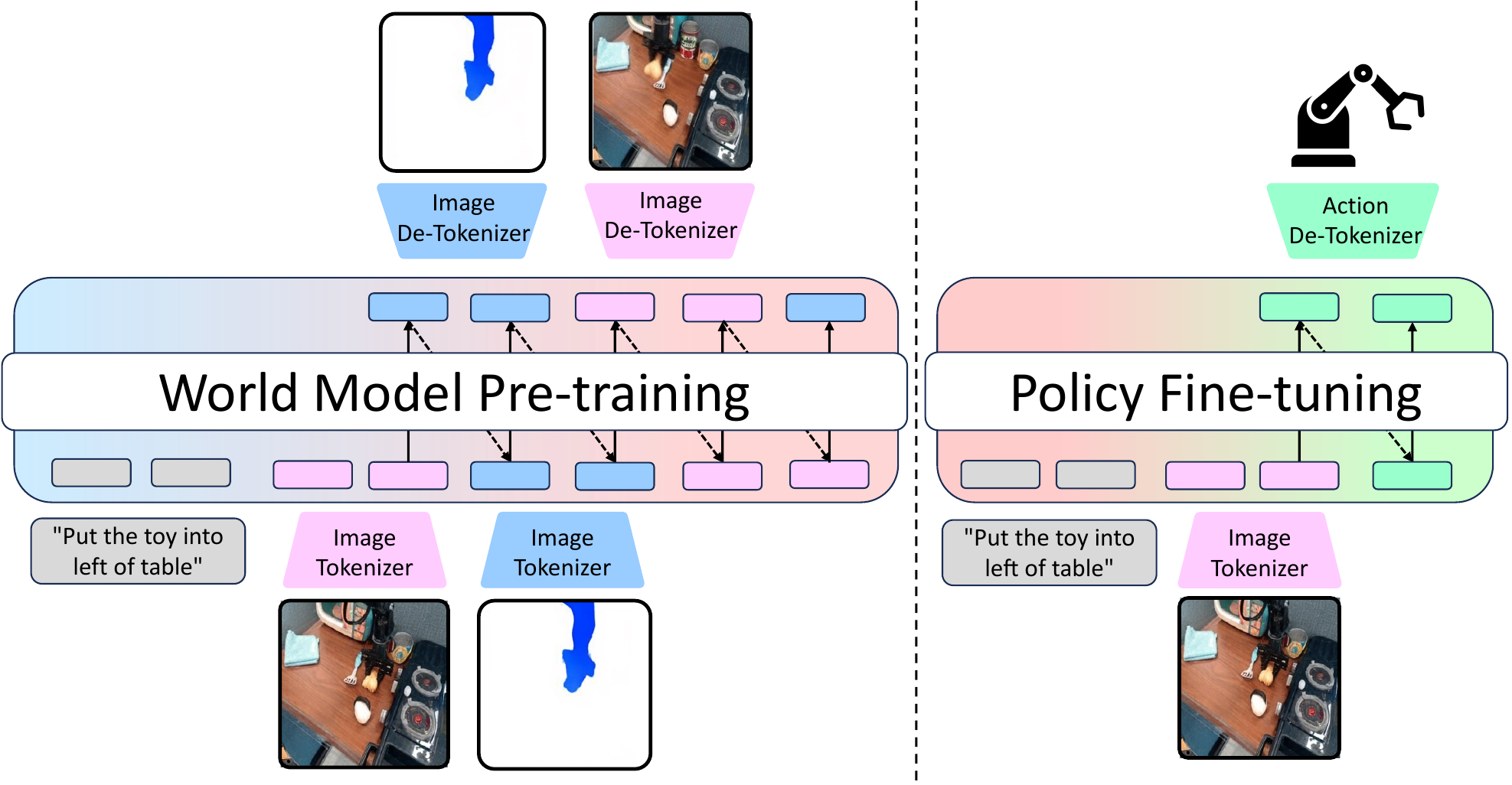}
    \caption{
    \textbf{Model Architecture of FlowVLA.} Our model instantiates the two-stage training paradigm in Figure~\ref{fig:teaser}.
    \textbf{(Left) Stage 1: World Model Pre-training with Visual CoT.} Input frames are encoded into appearance tokens (pink). The model then autoregressively predicts an interleaved sequence of motion tokens (blue) and future appearance tokens. Our proposed $v_t \rightarrow f_t \rightarrow v_{t+1}$ prediction forces the model to reason about dynamics before forecasting the future. For conceptual clarity, the Image and Flow Tokenizers are visualized separately; in practice, they are the exact same module applied to both appearance and motion inputs.
    \textbf{(Right) Stage 2: Policy Fine-tuning.} The pre-trained world model is adapted for action prediction. Conditioned on a text instruction (gray) and the current observation (magenta), the model autoregressively predicts action tokens (green) that are decoded into robot action chunk.
}
    \label{fig:framework}
\end{figure*}

Robotics manipulation in diverse and unstructured environments~\cite{10458370,10771693,11027446} has been a long-standing challenge, requiring both precise action prediction and robust understanding of visual and linguistic cues.
Recent advances in Vision-Language-Action (VLA) models~\cite{kim2024openvla,zitkovich2023rt,black2024pi0visionlanguageactionflowmodel,team2024octo}, particularly those pre-trained as world models like UniVLA~\cite{wang2025unifiedvisionlanguageactionmodel} and WorldVLA~\cite{cen2025worldvla}, have shown remarkable promise for creating generalist robots capable of tackling such manipulation tasks. The prevailing strategy involves training a large autoregressive transformer to predict the next visual frame given past observations, effectively learning the dynamics of the environment from vast amounts of video data. This learned world model then serves as a powerful foundation for fine-tuning downstream action policies.

Despite their success, these models suffer from a critical and foundational flaw: they attempt to predict the next frame\footnote{Throughout this paper, ``frame’’ denotes a sampled key frame from the video sequence rather than an immediately adjacent frame.} in a single, direct step, without explicitly considering the underlying physical motion. This next-frame prediction paradigm is often a ``pixel-copying trap''~\cite{ming2024survey}, where the model learns to replicate static backgrounds without a deep understanding of spatiotemporal dynamics, leading to blurry, inconsistent, and physically implausible long-horizon forecasts. Furthermore, this approach creates a significant domain gap between the passive, observational knowledge learned during pre-training and the active, action-oriented knowledge required for policy learning. This results in inefficient knowledge transfer and requires extensive fine-tuning, as evidenced by slow convergence on downstream tasks~\cite{zeng2024learning}.
 
One of the main reasons for the unsatisfactory performance of the above methods is that they attempt to learn a direct, unreasoned mapping from the current frame to the next, thereby bypassing the crucial step of physical reasoning.
Drawing inspiration from the success of Chain of Thought prompting in Large Language Models~\cite{wei2022chain}, which enhances reasoning by generating intermediate steps, we propose a novel paradigm for world models: a \textbf{Visual Chain of Thought (Visual CoT)}. Instead of a single, opaque leap from the current frame $v_t$ to the next $v_{t+1}$, we decompose the prediction into a structured reasoning process. First, predict the intermediate physical dynamics---the optical flow $f_t$ that describes \textit{how} every pixel will move. Then, conditioned on this explicit motion plan, predict the resulting future frame. This $v_t \rightarrow f_t \rightarrow v_{t+1}$ causal chain (see Figure~\ref{fig:teaser}, top) transforms the learning objective from mere pattern recognition into a structured physical reasoning task. By explicitly modeling dynamics, the world model learns representations that are inherently more aligned with the action-centric knowledge required for policy learning.

To fully leverage the above physically grounded world model for robot action prediction, we propose \textbf{FlowVLA}, a model that realizes the abstract dynamics step using optical flow as the concrete motion representation. Our training paradigm proceeds in \textbf{two stages} as illustrated in Figure~\ref{fig:teaser}.
\emph{Stage 1: World Model Pre-training with Visual CoT} focuses on visual reasoning, where the model is pre-trained on large-scale videos to master physically plausible and motion-coherent future frame forecasting. 
\emph{Stage 2: Policy Fine-tuning} adapts this pre-trained world model for action prediction: given a text instruction and current observation, the model now predicts discrete robot action chunk instead of future frames. 
Because Stage~1 has already aligned visual and dynamics representations with physical reality, Stage~2 can fine-tune the policy with substantially improved sample efficiency, directly benefiting from the explicit motion reasoning learned during World Model Pre-training with Visual CoT.

A key aspect of our design is to integrate motion without introducing dedicated architectural components. We encode optical flow fields by a flow map, allowing them to be processed by the exact same Vector Quantization (VQ) tokenizer as regular camera observations. This enables a single, unmodified autoregressive transformer to seamlessly learn the interleaved sequence of appearance and motion tokens. This design makes FlowVLA a truly Unified Visual CoT, where the reasoning steps (flows) and states (frames) are expressed in a shared vocabulary and processed by a single, unified model.

Our work makes the following contributions:
\begin{itemize}
    \item We identify a fundamental limitation of next-frame prediction and propose Visual Chain of Thought (Visual CoT) as a new paradigm for learning dynamics for VLA world models, explicitly modeling motion to better bridge the gap between world model pre-training and policy fine-tuning.
    \item We introduce FlowVLA, an effective instantiation of this paradigm that unifies appearance and motion reasoning within a single autoregressive Transformer via shared tokenization, avoiding task-specific architectural components and enabling seamless integration of motion reasoning into world model pre-training.
    \item We demonstrate through extensive experiments that FlowVLA achieves state-of-the-art performance on challenging manipulation benchmarks and a real-robot platform, while offering superior sample efficiency.
\end{itemize}

\vspace{-1mm}
\section{Methodology}
\label{sec:method}
In this section, we introduce FlowVLA, a novel framework designed to instantiate our proposed Visual Chain of Thought (Visual CoT) paradigm. Firstly, we present our Visual CoT formulation. Secondly, We provide a high-level overview of our two-stage training. Thirdly, we detail the Visual CoT pre-training stage, including our unified tokenization scheme for appearance and motion. Finally, we describe how the learned world model is finetuned for downstream robotics tasks.

\subsection{Visual Chain of Thought}
\label{subsubsec:reframing}
The paradigm for pre-training world models is next-frame prediction. This approach aims to learn a probabilistic model, typically parameterized by a large Transformer parameterized by $\theta$, that predicts the next visual observation $v_{t+1}$ given a history of past observations and a guiding language instruction $L$. The learning objective can be expressed as maximizing the log-likelihood of the next frame:
\begin{equation}
    \max_\theta \mathbb{E}_{(v_t, v_{t+1}, L) \sim \mathcal{D}} [\log P_\theta(v_{t+1} | v_t, L)].
\label{eq:ntp}
\end{equation}
While conceptually simple and scalable to large and unlabeled video datasets $\mathcal{D}$, this formulation suffers from fundamental limitations that hinder the acquisition of robust physical reasoning.

To overcome the aforementioned challenges, we reframe the world modeling task by introducing a \textbf{Visual Chain of Thought (Visual CoT)}. Inspired by the success of CoT in large language models~\cite{wei2022chain}, which improves reasoning by generating intermediate steps, we propose to decompose the visual prediction process. Instead of a single, unreasoned leap, we compel the model to first ``think'' about the intermediate physical process before generating the final outcome. We instantiate this ``thought'' as the dense optical flow field $f_t$, which describes the per-pixel motion from $v_t$ to $v_{t+1}$.

Formally, we reformulate the task from modeling $P(v_{t+1} | v_t, L)$ in Equation~(\ref{eq:ntp}) to modeling the joint probability $P(v_{t+1}, f_t | v_t, L)$. By applying the chain rule of probability, we factorize this joint probability into a causal sequence:
\begin{equation}
P(v_{t+1}, f_t | v_t, L) = \underbrace{P(v_{t+1} | f_t, v_t, L)}_{\text{Appearance Generation}} \times \underbrace{P(f_t | v_t, L)}_{\text{Motion Reasoning}}.
\label{eq:visual_cot}
\end{equation}
This decomposition offers several key advantages. It decouples the learning problem, allowing the model to first focus on the physically-grounded \textit{Motion Reasoning} task before tackling the more appearance-focused \textit{Appearance Generation} task. It introduces a powerful \textbf{inductive bias}, explicitly guiding the model to learn a representation of motion, thereby grounding its predictions in physical causality. Crucially, this \textbf{aligns} the pre-training objective with the needs of downstream action prediction. A model that explicitly understands motion $f_t$ is inherently better prepared to generate an action chunk $a_t$ that cause desired motions.

The above reformulation addresses the limitations of the direct next‑frame prediction paradigm.
From a learning perspective, the objective function in Equation~(\ref{eq:ntp}) is fundamentally ill-posed. It frames world modeling as a high-dimensional regression problem directly in pixel space, which creates an optimization landscape fraught with trivial local minima. This often manifests as the ``pixel-copying trap''~\cite{ming2024survey}, where the model discovers that the easiest way to minimize reconstruction error is to simply replicate static background pixels from the input frame. This optimization shortcut is a primary cause of the blurry, inconsistent, and physically implausible forecasts observed in models trained with this paradigm. Furthermore, this direct mapping $v_t \rightarrow v_{t+1}$ lacks an explicit \textbf{causal structure}. The model learns a direct correlation between pixel configurations over time, which may not correspond to the true physical causal relationships. This reliance on correlation over causation results in sensitive models that fail to generalize to out-of-distribution scenarios where visual cues change, even if the underlying physics remain the same.

\subsection{Framework Overview}

As shown in Figure~\ref{fig:framework}, FlowVLA follows a two-stage training paradigm, consistent with state-of-the-art methods like UniVLA~\cite{wang2025unifiedvisionlanguageactionmodel} and WorldVLA~\cite{cen2025worldvla} to ensure a fair basis for comparison.
\begin{enumerate}
    \item Stage 1: World Model Pre-training: The model learns general physical dynamics from large-scale, action-free video data by executing our proposed Visual Chain of Thought.
    \item Stage 2: Policy Finetuning: The pre-trained model weights are finetuned on downstream, action-annotated robotics datasets to learn specific control policies.
\end{enumerate}

\subsection{Stage 1: World Model Pre-training via Visual Chain of Thought}
The goal of this stage is to learn a robust world model by compelling it to reason about dynamics before predicting future states. This is achieved through our Visual Chain of Thought (Visual CoT) pre-training task. Below, we detail the tokenization scheme that unifies appearance and motion, and then describe the autoregressive objective used to learn the reasoning chain.

\paragraph{Unified Motion and Appearance Tokenization}
A cornerstone of FlowVLA's design is a unified tokenization scheme that represents two physically distinct signals—appearance (images) and motion (optical flow)—within a single, shared vocabulary. This approach preserves architectural simplicity and promotes the learning of deep cross-modal representations. To achieve this, we process each modality as follows.

For the \textbf{Appearance Representation}, standard RGB frames $v_t$, which capture the static appearance of the scene, are processed by a pretrained VQ-GAN tokenizer~\cite{esser2021taming}. This tokenizer discretizes each high-resolution image into a compact grid of visual tokens from a learned codebook. This tokenized representation allows the Transformer to process complex visual information in the same sequential format as text.

For the \textbf{Motion Representation}, we encode the abstract motion dynamics using optical flow $f_t$. It is a dense pixel-level representation that describes the projected motion of every point in the visual field between two consecutive frames. We choose optical flow for its ability to capture fine-grained interaction dynamics, such as sliding, pushing, and rotating. Another advantage is the availability of robust off-the-shelf models, such as RAFT~\cite{teed2020raft}, for pre-computation from video data.

We choice optical flow instead of object-centric alternatives like 3D poses or bounding boxes for two main reasons. First, acquiring such object-level representations accurately often requires specialized upstream models trained on large, manually annotated datasets, limiting scalability and introducing potential points of failure. Second, these sparse representations cannot capture non-rigid motion or complex interaction dynamics. In contrast, optical flow provides a dense, general signal that is independent of object detectors. Therefore, it naturally represents continuous motion.

Critically, the image structure of optical flow allows for a shared tokenizer with RGB frames, ensuring tight modality alignment and architectural simplicity. These ``flow maps'' are then processed by the exact same VQ-GAN tokenizer used for the RGB frames. This design is central to our framework's efficiency and simplicity, yielding three significant benefits. First, it provides \textbf{Parameter Efficiency}, as no new motion-specific tokenizer or architectural modules are required. Second, it maintains \textbf{Architectural Simplicity} through a single, end-to-end autoregressive pipeline without specialized branches. Finally, it fosters a \textbf{Unified Representation}, encouraging the model to learn deep correlations between appearance (``what is there'') and motion (``how it moves'') within a shared latent space.

To integrate optical flow into our unified framework, we convert the 2-channel flow fields (containing displacements $u$ and $v$) into standard 3-channel RGB images. This conversion, following the technique from VideoJAM~\cite{chefer2025videojam}, maps the motion vector at each pixel to a color based on its polar coordinates. 
Unlike traditional optical flow visualizations that often normalize motion magnitude by the global maximum displacement in a frame, causing subtle motions to be visually suppressed, VideoJAM applies a fixed scaling coefficient and non-linear normalization strategy, which preserves fine-grained motion cues while avoiding saturation in high-speed regions.
The direction of motion is mapped to the color's Hue (from angle $\alpha=\arctan 2(v,u)$), and the speed of motion is mapped to the color's Saturation and Value (from magnitude $m=\sqrt{u^{2}+v^{2}}$). 
To handle a wide range of motion speeds without saturation or loss of detail for subtle movements, the magnitude is non-linearly normalized to the range $[0,1]$ using a scaling coefficient $\sigma=0.15$:
\begin{equation}
m_{\text{norm}}=\min\left(1.0,\frac{m}{\sigma \cdot \sqrt{H^{2}+W^{2}}}\right),
\end{equation}
where $H$ and $W$ are the frame's height and width.

\paragraph{Autoregressive Learning of the Visual CoT}

With a unified token representation for both frames ($v_t$) and flow ($f_t$), we construct a reasoning chain $v_t \rightarrow f_t \rightarrow v_{t+1}$. We employ a standard decoder-only Transformer, training it to predict an interleaved sequence of frames and optical flow fields given an optional language instruction $L_{\text{instr}}$:
\begin{align}
    S_{\text{wm}} = \{L_{\text{instr}}, v_0, f_0, v_1, f_1, \dots, v_T, f_T \}
    \label{eq:sequence_simple}
\end{align}
The model is trained using a standard next-token prediction objective, maximizing the log-likelihood of the sequence. The loss of the world model, $\mathcal{L}_{\text{WM}}$, is the sum of the cross-entropy losses in both the reasoning step (flow tokens) and the final state (next frame tokens). Formally, for each timestep $t$, the model first predicts the flow $f_t$ based on all preceding tokens, and then predicts the next frame $v_{t+1}$ conditioned on both the history and the just-predicted flow:
\begin{equation}
    \mathcal{L}_{\text{WM}} = \sum_{t=0}^{T-1} \left( \mathcal{L}_{\text{CE}}(f_t | S_{< v_{t+1}}) + \lambda \cdot \mathcal{L}_{\text{CE}}(v_{t+1} | S_{< v_{t+1}}, f_t) \right)
    \label{eq:loss_wm}
\end{equation}
where $S_{<v_{t+1}}$ denotes all the tokens preceding $v_{t+1}$, and $\lambda$ is a balancing hyperparameter (set to 1.0 in our experiments). This objective explicitly forces the model to perform a ``reason $\rightarrow$ predict'' process during both training and inference.

\subsection{Stage 2: Finetuning for Action Prediction}

\paragraph{Initialization and Task.} The policy model is initialized with the weights from the pre-trained world model. During this stage, the input sequence is composed of interleaved observations and actions: $S_{\text{policy}} = \{L_{\text{instr}}, v_0, a_0, v_1, a_1, \dots\}$, where $a_t$ represents the robot's action tokens.

\paragraph{Action Tokenization and Objective.} Actions are discretized into tokens following the FAST~\cite{pertsch2025fast}. Critically, the fine-tuning loss, $\mathcal{L}_{\text{policy}}$, is computed only over the action tokens. This objective directs the model to leverage all its learned visual and dynamical knowledge towards the singular goal of making correct action prediction.

\vspace{-2mm}
\section{Experiments}

We conduct a comprehensive set of experiments to validate the effectiveness of our proposed Visual Chain of Thought framework. Our evaluation is designed to answer four key questions:
\begin{enumerate}
    \item[\textbf{Q1:}] Does FlowVLA achieve state-of-the-art performance on complex, long-horizon robotics tasks?
    \item[\textbf{Q2:}] Does explicit motion reasoning lead to superior world modeling capabilities compared with approaches that learn world dynamics implicitly?
    \item[\textbf{Q3:}] Is FlowVLA more sample-efficient during policy fine-tuning, validating our claim of bridging the pre-training/fine-tuning gap?
    \item[\textbf{Q4:}] What is the effectiveness of each key component in our model architecture?
\end{enumerate}

\subsection{Experimental Setup}
To comprehensively assess FlowVLA's capabilities, we conduct evaluations across a suite of complementary settings. We use two challenging simulation benchmarks, LIBERO and SimplerEnv, to measure generalization and robustness against domain shifts. Additionally, we perform Real-Robot experiments to validate the model's practical applicability and its ability to transfer skills from simulation to reality.

\textbf{LIBERO Benchmark.} We evaluate FlowVLA on the LIBERO benchmark~\cite{liu2023libero} to assess its generalization across multiple axes. Following the standard behavioral cloning setup, we report performance on its four main suites, which test for generalization to novel spatial layouts, objects, task goals, and long-horizon compositional challenges.

\textbf{SimplerEnv Benchmark.} We use SimplerEnv~\cite{li2024evaluating} to assess the model's robustness against significant domain shifts. This benchmark is specifically designed to evaluate policy transfer by introducing diverse variations in lighting, textures, and camera viewpoints, which are more representative of real-world complexity.

\begin{figure*}[t]
    \centering
    \begin{subfigure}{0.44\textwidth}
        \centering
        \includegraphics[width=\linewidth]{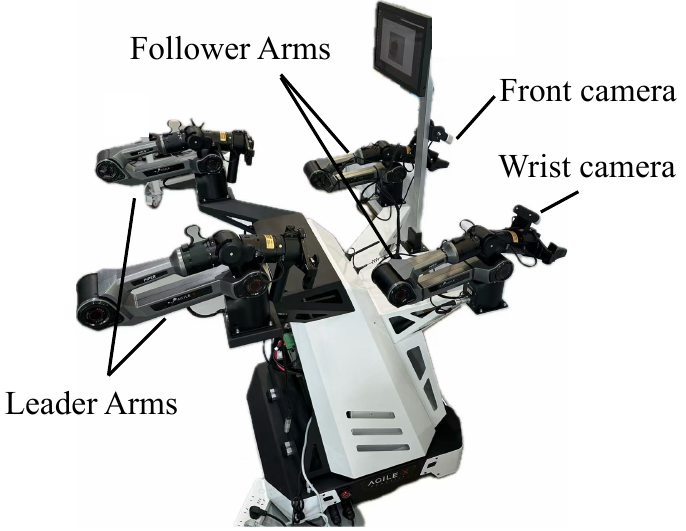}
        \caption{\textbf{Platform setup}}
        \label{fig:cobot_platform}
    \end{subfigure}
    \hspace{0.04\textwidth} 
    \begin{subfigure}{0.44\textwidth}
        \centering
        \includegraphics[width=\linewidth]{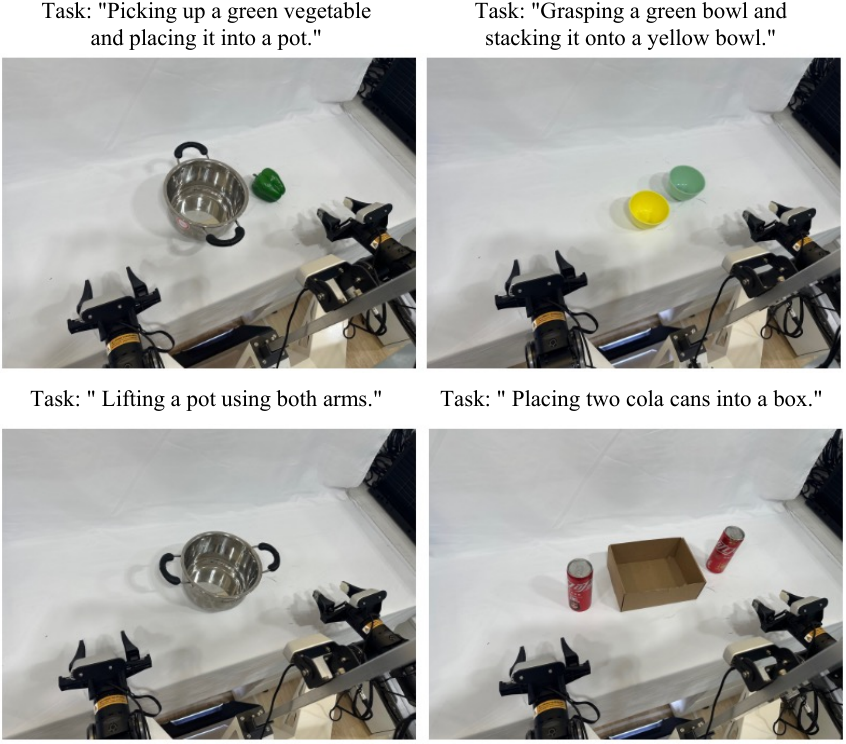}
        \caption{\textbf{Example tasks}}
        \label{fig:real_world}
    \end{subfigure}

    \caption{
        AgileX Cobot dual-arm platform and real-world manipulation tasks: (a) system setup: leader arms are user-operated, follower arms mirror the actions. Vision is provided by a front camera for global scene view and wrist cameras for close-up workspace observation. (b) representative single-arm and bimanual operations: from simple single-arm tasks to complex long-horizon bimanual manipulations.
    }
    \label{fig:cobot_combined}
\end{figure*}

\textbf{Real-world Experiments based on AgileX Cobot.}
As shown in Figure~\ref{fig:cobot_combined}(a), we conduct our real-world experiments on a Cobot dual-arm robot manufactured by AgileX Robotics, which adopts the Mobile ALOHA system design~\cite{fu2024mobile}. Each arm has 7 degrees of freedom and is equipped with a parallel gripper. The robot carries multiple onboard sensors, including wrist-mounted cameras on both arms and a front-facing camera, for capturing RGB observations used as model input.
We design four manipulation tasks to comprehensively evaluate the model’s spatial reasoning and control capabilities (see Figure~\ref{fig:cobot_combined}(b)). For each task, we collect 50–200 human-teleoperated demonstrations for fine-tuning, thereby evaluating the model’s data efficiency and its ability to rapidly adapt to the specific embodiment of the Cobot platform.

\paragraph{Implementation Details.}
Our FlowVLA model is built on the 8.5B parameter Emu3~\cite{wang2024emu3nexttokenpredictionneed} and UniVLA~\cite{wang2025unifiedvisionlanguageactionmodel} architecture. We incorporate optical flow, pre-computed with RAFT~\cite{teed2020raft}, as an additional modality to represent motion. 
We follow the standard training setup for the LIBERO and SimplerEnv benchmarks~\cite{kim2024openvla,kim2025finetuningvisionlanguageactionmodelsoptimizing}.
For LIBERO, we pre-train the world model for 5k steps with a batch size of 16, and then fine-tune the policy for 5k steps with a batch size of 96. For the SimplerEnv benchmark, pre-training runs for 12k steps with a batch size of 32 and policy fine-tuning for 20k steps with a batch size of 128.

\subsection{Evaluations Results (Q1)}
To answer Q1, we evaluate the final performance of FlowVLA after policy finetuning on both benchmarks. Our method establishes a new state-of-the-art on both, demonstrating its effectiveness and robustness.

\textbf{Results on LIBERO.} As shown in Table~\ref{tab:libero_results}, FlowVLA consistently outperforms all prior methods across the four evaluated suites. Notably, the performance gains are most significant on the \textit{Long} horizon tasks. This directly highlights the benefit of learning a world model with a more robust understanding of physical dynamics, as our Visual CoT framework enables better long-term planning and reasoning.

\textbf{Results on SimplerEnv.} We further test our model's robustness on the SimplerEnv benchmark, which introduces significant visual domain shifts. Table~\ref{tab:simplerenv_results} shows that FlowVLA achieves a substantial improvement over existing methods. The remarkable success on tasks that were previously challenging for other models (e.g., stacking blocks) validates that our explicit motion reasoning leads to policies that are more resilient to the visual and physical variations found in more realistic environments.

\textbf{Results on Real-Robot.}
As shown in Table~\ref{tab:cobot_realworld_results}, FlowVLA exhibits clear advantages over all baselines across the four real-world Cobot tasks, including both single-arm and bimanual operations. The improvements are particularly pronounced in more complex bimanual tasks such as \textit{Placing two cola cans into a box} and \textit{Lifting a pot using both arms}, where precise coordination and dynamic interaction are required. During evaluation, each task was executed 25 times to ensure statistically reliable success rates. These results confirm that FlowVLA’s explicit motion reasoning and world model-based long-horizon planning are effective in real-world settings, enabling rapid adaptation to the specific embodiment and visual conditions of the Cobot platform.

\begin{table}[t!]
\centering
\caption{
    \textbf{Results on the LIBERO Benchmark~\cite{liu2023libero}.} 
    We report the final task success rate (\%) and compare FlowVLA against state-of-the-art methods, grouped by their core methodology. The results demonstrate that our Visual CoT pre-training leads to superior performance, highlighting the efficiency of our proposed framework.
}
\label{tab:libero_results}

\setlength{\tabcolsep}{3pt} 

\begin{tabularx}{\textwidth}{ >{\raggedright\arraybackslash}p{5.8cm} c *{5}{>{\centering\arraybackslash}X} }
\toprule
Model & \makecell{Large Scale\\Pretrain} & {Spatial} & {Object} & {Goal} & {Long} & {\textbf{Avg.}} \\
\midrule

\multicolumn{7}{@{}l}{\textit{\textbf{w/o World Model}}} \\
Diffusion Policy~\cite{chi2023diffusion} & $\times$      & 78.3 & 92.5 & 68.3 & 50.5 & 72.4 \\
Octo~\cite{team2024octo}                 & $\checkmark$  & 78.9 & 85.7 & 84.6 & 51.1 & 75.1 \\
OpenVLA~\cite{kim2024openvla}            & $\checkmark$  & 84.7 & 88.4 & 79.2 & 53.7 & 76.5 \\
DiT Policy~\cite{hou2025dita}      & $\checkmark$  & 84.2 & 96.3 & 85.4 & 63.8 & 82.4 \\
TraceVLA~\cite{zheng2024tracevla}  & $\checkmark$  & 84.6 & 85.2 & 75.1 & 54.1 & 74.8 \\
SpatialVLA~\cite{qu2025spatialvla}  & $\checkmark$  & 88.2 & 89.9 & 78.6 & 55.5 & 78.1 \\
pi0-FAST~\cite{pertsch2025fast}  & $\checkmark$  & 96.4 & 96.8 & 88.6 & 60.2 & 85.5 \\
ThinkAct~\cite{huang2025thinkact} & $\checkmark$  & 88.3 & 91.4 & 87.1 & 70.9 & 84.4 \\
\midrule
\multicolumn{7}{@{}l}{\textit{\textbf{w/ World Model}}} \\
WorldVLA~\cite{cen2025worldvla}       & $\times$      & 85.6 & 89.0 & 82.6 & 59.0 & 79.1 \\
UniVLA$^\dag$~\cite{wang2025unifiedvisionlanguageactionmodel}    & $\times$      & {92.6} & {93.8} & {86.6} & 63.0 & {84.0} \\
CoT-VLA~\cite{zhao2025cot}          & $\checkmark$  & 87.5 & 91.6 & 87.6 & 69.0 & 81.1 \\
\rowcolor{gray!20}
\textbf{FlowVLA (ours)}                 & $\times$      & \bfseries 93.2 & \bfseries 95.0 & \bfseries 91.6 & \bfseries 72.6 & \bfseries \textbf{88.1} \\

\bottomrule
\end{tabularx}

\begin{minipage}{\textwidth}
\vspace{2pt} 
\footnotesize 
\textdagger\ Our reported UniVLA result is from our re-implementation, pre-trained only on LIBERO without wrist camera images for a fair comparison.
\end{minipage}

\end{table}

\begin{table}[t!]
\centering
\caption{
    \textbf{Results on the SimplerEnv-WidowX benchmark~\cite{li2024evaluating}.} We report the final task success rate (\%). FlowVLA significantly surpasses prior methods, demonstrating superior robustness to the visual domain shifts present in this benchmark. Best results are in \textbf{bold}.
}
\label{tab:simplerenv_results}
\resizebox{\textwidth}{!}{%
\begin{tabular}{@{}lccccc@{}}
\toprule
Model & Put Spoon & Put Carrot & Stack Block & Put Eggplant & \textbf{Avg.} \\
\midrule
RT-1-X~\cite{team2024octo} & 0.0 & 4.2 & 0.0 & 0.0 & 1.1 \\
Octo-Base~\cite{team2024octo} & 12.5 & 8.3 & 0.0 & 43.1 & 16.0 \\
Octo-Small~\cite{team2024octo} & 47.2 & 9.7 & 4.2 & 56.9 & 29.5 \\
OpenVLA~\cite{team2024octo} & 0.0 & 0.0 & 0.0 & 4.1 & 1.0 \\
RoboVLMs~\cite{liu2025towards} & 45.8 & 20.8 & 4.2 & 79.2 & 37.5 \\
SpatialVLA~\cite{qu2025spatialvla} & 16.7 & 25.0 & 29.2 & 100 & 42.7 \\
RoboPoint~\cite{yuan2024robopoint} & 16.7 & 20.8 & 8.3 & 25.0 & 17.7 \\
FSD~\cite{yuan2025seeing} & 41.6 & 50.0 & 33.3 & 37.5 & 40.6 \\
Embodied-R1~\cite{yuan2025embodiedr1reinforcedembodiedreasoning} & 62.5 & \textbf{68.0} & 36.1 & 58.3 & 56.2 \\
ThinkAct~\cite{huang2025thinkact} & 58.3 & 37.5 &  8.7 & 70.8 & 43.8 \\
UniVLA$^\dag$~\cite{wang2025unifiedvisionlanguageactionmodel} & 62.5 & 62.5 & 41.6 & 95.8 & 65.6 \\
\rowcolor{gray!20}
\textbf{FlowVLA (Ours)} & \textbf{70.8} & 62.5 & \textbf{62.5} & \textbf{100.0} & \textbf{74.0} \\
\bottomrule
\end{tabular}%
}
\begin{minipage}{\textwidth}
\vspace{2pt} 
\footnotesize 
\textdagger\ Result obtained by evaluating the officially released checkpoint.
\end{minipage}
\end{table}

\begin{table}[t!]
\centering
\caption{
    \textbf{Results on the Real-world AgileX Cobot platform.} We report the task success rates (\%). The evaluation covers four manipulation tasks of varying difficulty, from single-arm to bimanual operation. Best results are in \textbf{bold}.
}
\label{tab:cobot_realworld_results}
\resizebox{\textwidth}{!}{%
\begin{tabular}{@{}lccccc@{}}
\toprule
Model & Stack Bowls & Place Vegetable & Place Bottles & Lift Pot & \textbf{Avg.} \\
\midrule
ACT~\cite{zhao2023learning} & 32.0 & 24.0 & 12.0 & 8.0 & 19.0 \\
OpenVLA~\cite{kim2024openvla} & 28.0 & 20.0 & 20.0 & 12.0 & 20.0 \\
UniVLA~\cite{wang2025unifiedvisionlanguageactionmodel} & 48.0 & 40.0 & 16.0 & 20.0 & 31.0 \\
\rowcolor{gray!20}
\textbf{FlowVLA (Ours)} & \textbf{56.0} & 60.0 & \textbf{32.0} & \textbf{28.0} & \textbf{44.0} \\
\bottomrule
\end{tabular}%
}
\begin{minipage}{\textwidth}
\vspace{2pt} 
\end{minipage}
\end{table}

\subsection{Analysis of World Modeling Capabilities (Q2)}
\begin{figure*}[!h]
    \centering
    
    \begin{subfigure}{\textwidth}
        \includegraphics[width=0.95\textwidth]{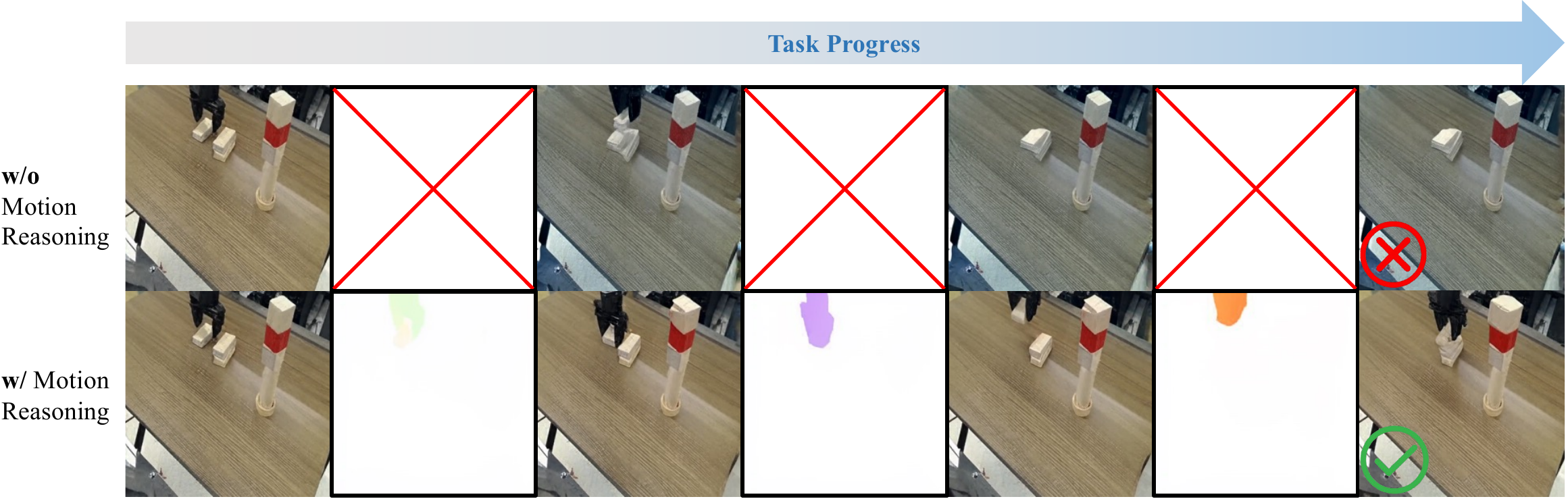} 
        \caption{Task: "Put the rectangular on top of the rectangular block next to it."}
        \label{subfig:task_a_phys}
    \end{subfigure}
    
    \vspace{3mm} 
    
    \begin{subfigure}{\textwidth}
        \includegraphics[width=0.95\textwidth]{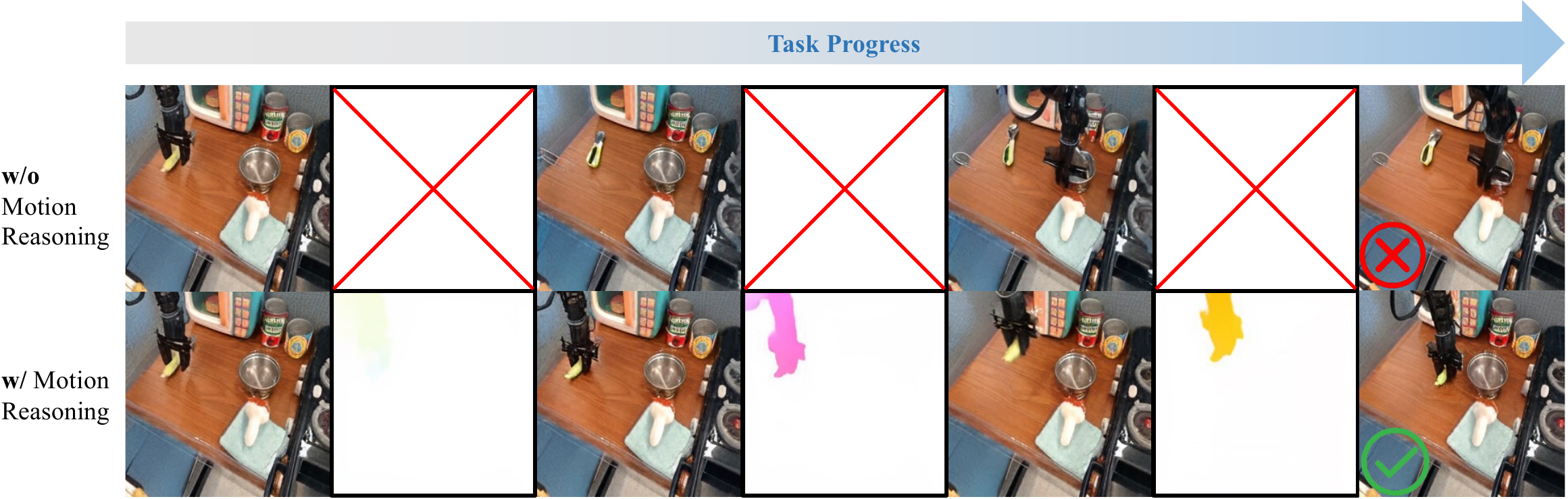} 
        \caption{Task: "Move the spoon so that it sits to the left of the metal pot."}
        \label{subfig:task_b_phys}
    \end{subfigure}
    
    \caption{
        \textbf{Analysis of Physical Plausibility on the Bridge V2 Dataset.} 
        This figure highlights common physical failures in the baseline model. In both examples, the baseline model (top row) struggles to maintain physical consistency, leading to implausible outcomes such as a disappearing manipulator or erratic object behavior. In contrast, \textbf{FlowVLA} (bottom row), guided by its motion-first reasoning, produces stable and physically coherent predictions that accurately reflect the scene's dynamics.
    }
    \label{fig:world_model_viz_bridge_1}
\end{figure*}

\begin{figure*}[!h]
    \centering
    \begin{subfigure}{\textwidth}
        \includegraphics[width=0.95\textwidth]{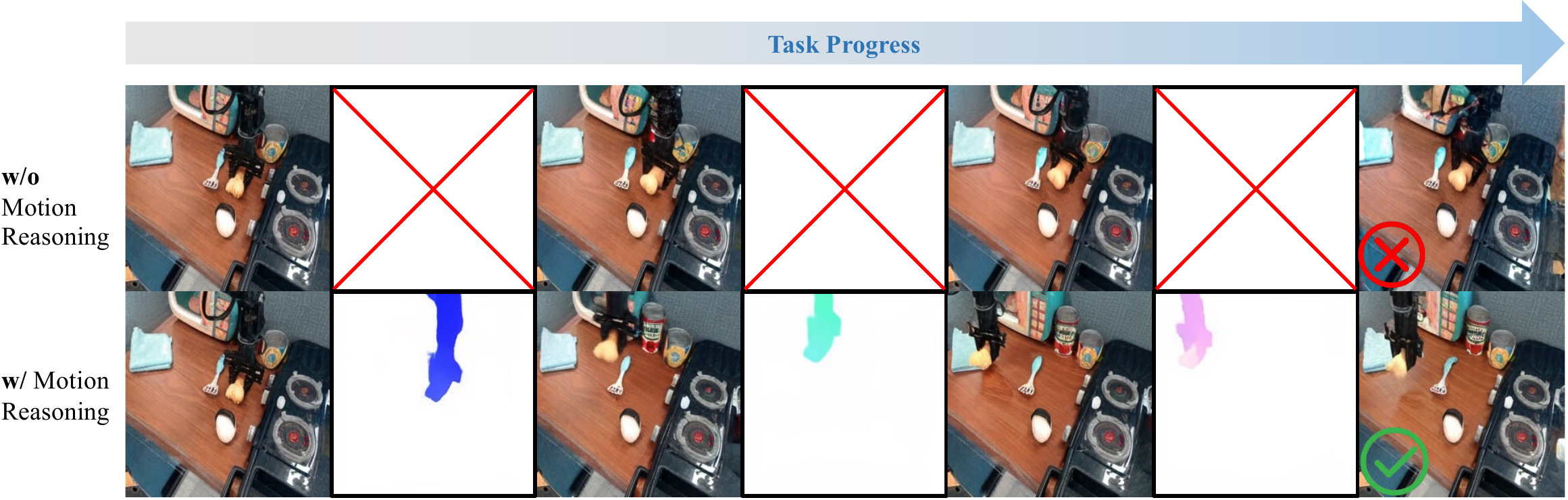} 
        \caption{Task: "Put the toy into left of table."}
        \label{subfig:task_a_sem}
    \end{subfigure}
    
    \vspace{3mm} 
    
    \begin{subfigure}{\textwidth}
        \includegraphics[width=0.95\textwidth]{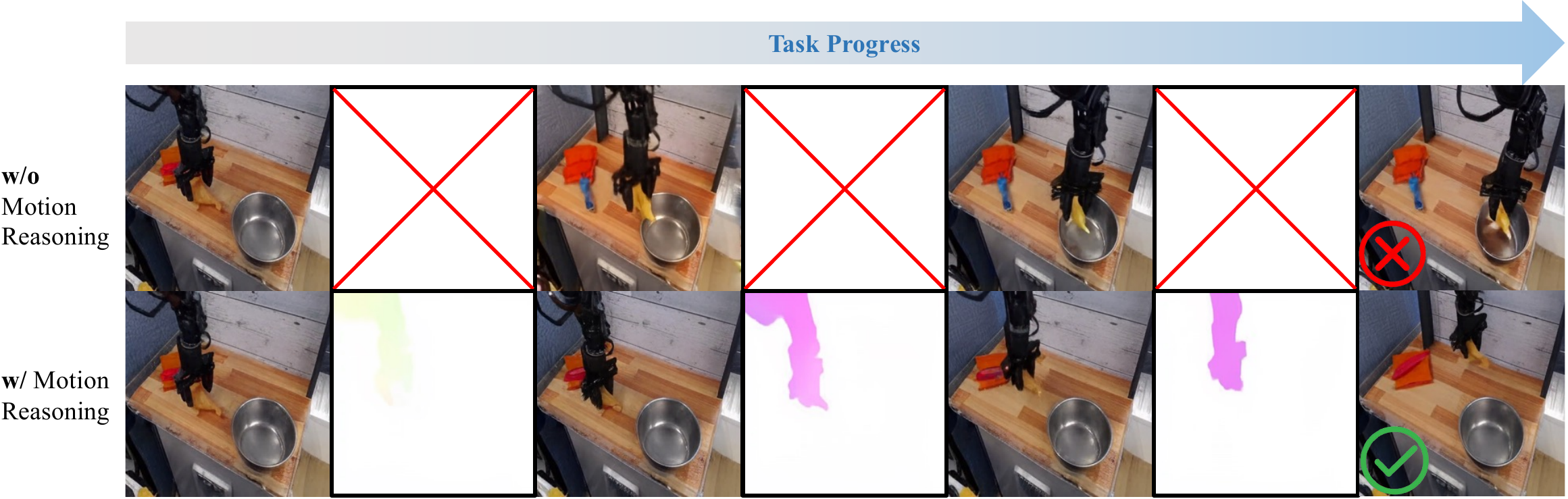} 
        \caption{Task: "Move toy diagonally little bit top on the right side."}
        \label{subfig:task_b_sem}
    \end{subfigure}
    
    \caption{
        \textbf{Analysis of Semantic Alignment on the Bridge V2 Dataset.} 
        This figure illustrates the baseline's failure to align predictions with language instructions. While the predicted frames from baseline model (top row) might appear visually plausible at a glance, the resulting motion does not correspond to the specified task (e.g., moving an object in the wrong direction). \textbf{FlowVLA} (bottom row) again demonstrates superior performance, correctly interpreting the command and generating a corresponding visual trajectory. This underscores that our Visual CoT not only improves physical realism but also enhances the model's ability to ground language in action.
    }
    \label{fig:world_model_viz_bridge_2}
\end{figure*}

To demonstrate the superiority of our motion reasoning framework compared with approaches that learn world dynamics implicitly, we conduct a detailed qualitative analysis on the challenging, real-world Bridge V2 dataset. 
The standard next-frame prediction baseline suffers from two distinct and critical failure modes.
\begin{itemize}
    \item \textbf{Failures in Physical Plausibility.} As highlighted in Figure~\ref{fig:world_model_viz_bridge_1}, the baseline model generates physically incoherent rollouts, such as causing the robotic arm to suddenly vanish or producing inconsistent object motion. This indicates a fundamental inability to model the basic physical continuity of a scene.

    \item \textbf{Semantic Inconsistency.} Figure~\ref{fig:world_model_viz_bridge_2} illustrates a more subtle but equally critical issue. While the predicted frames from the baseline may appear visually coherent, the depicted action fails to follow the given language command. This reveals a disconnection between language understanding and visual forecasting.
\end{itemize}

In contrast, FlowVLA successfully overcomes the above challenges. By first reasoning about motion dynamics via optical flow, our model generates predictions that are not only physically plausible but also semantically aligned with the task instructions, showcasing the robustness and generalizability of our approach.

\subsection{Convergence Speed and Data Efficiency(Q3)}
\begin{figure}[h!]
    \centering
    
    \begin{subfigure}[b]{0.48\textwidth}  
        \includegraphics[width=\textwidth]{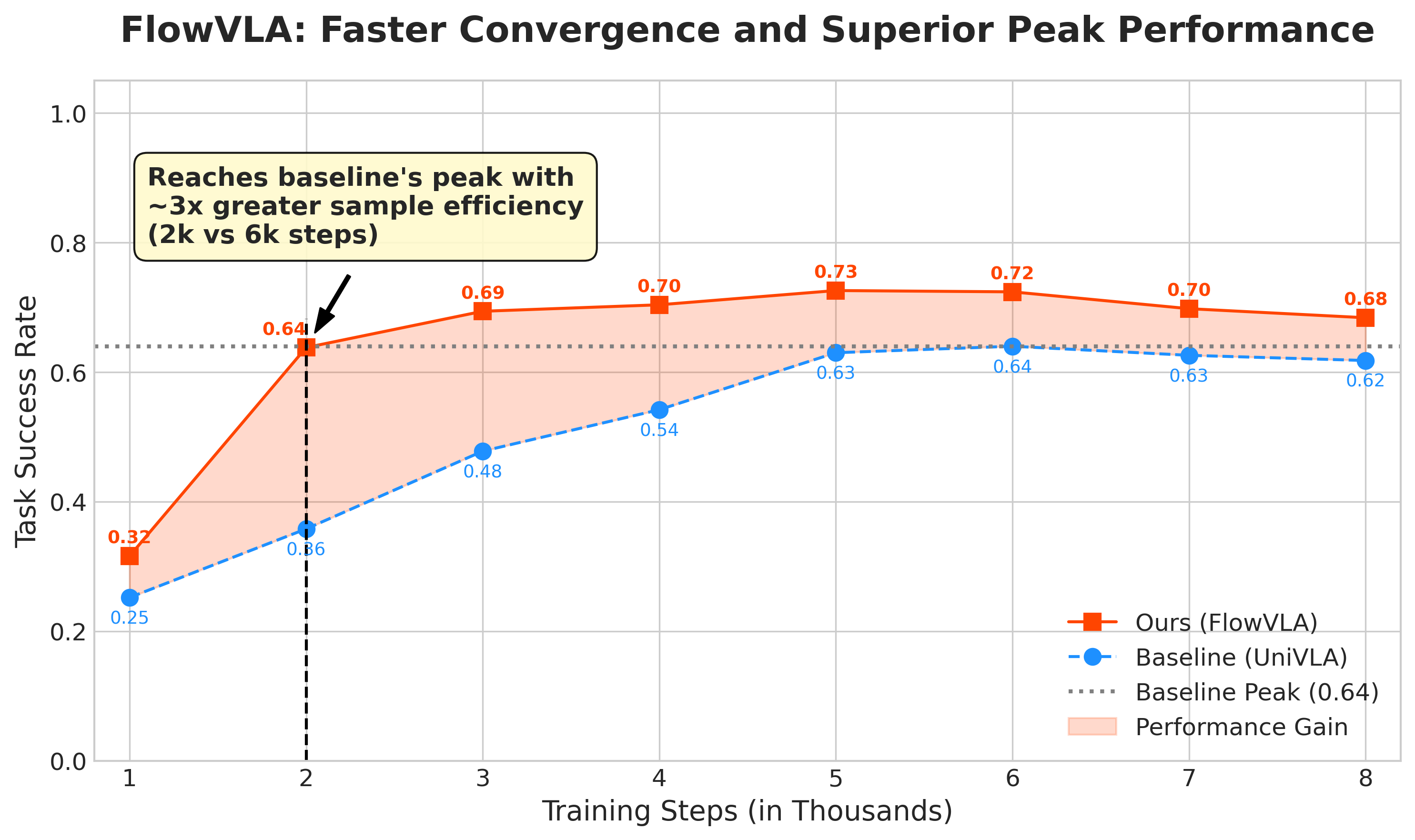}
        \caption{Training on 100\% of data}
        \label{fig:convergence_full}
    \end{subfigure}
    \hfill 
    \begin{subfigure}[b]{0.48\textwidth}  
        \includegraphics[width=\textwidth]
        {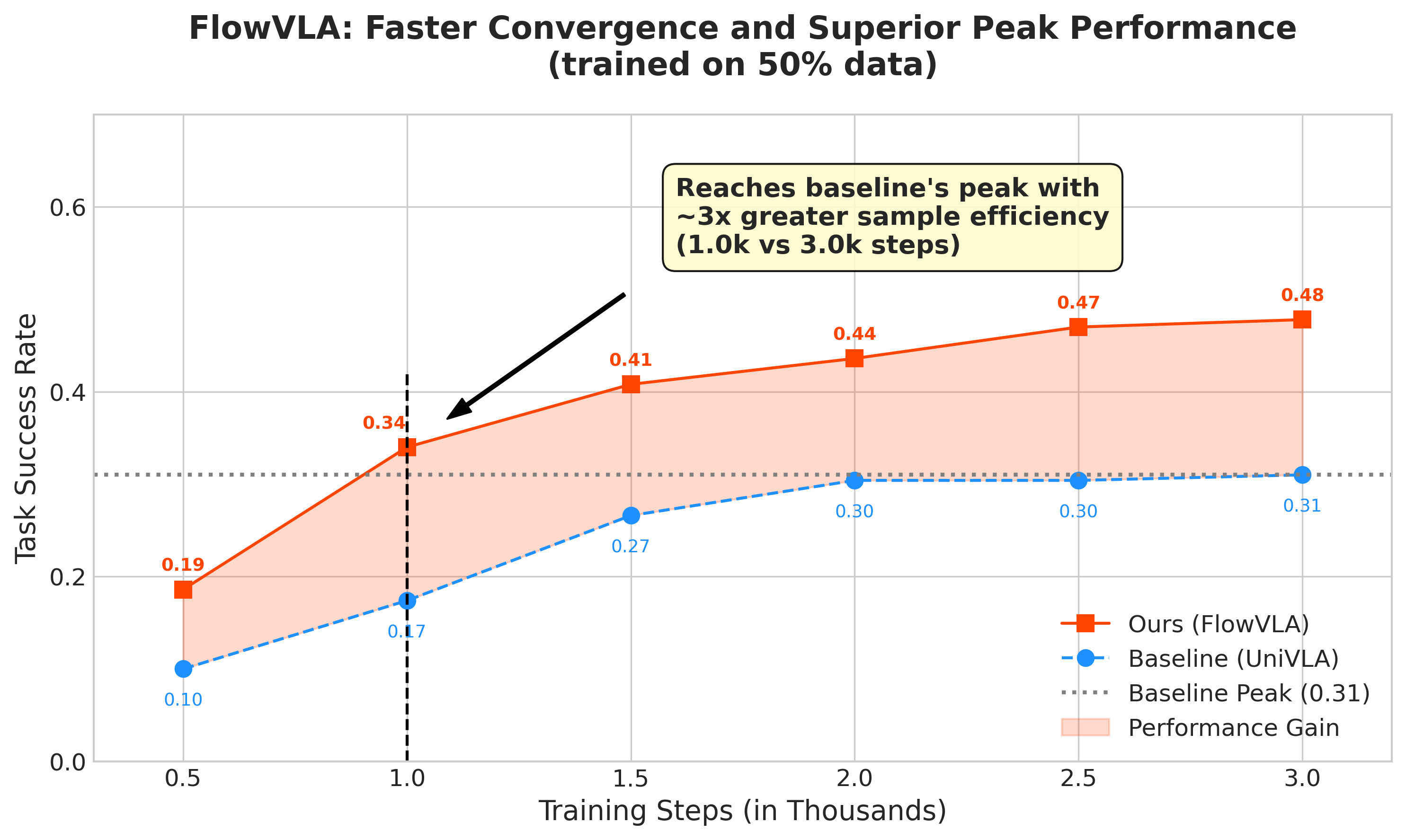} 
        \caption{Training on 50\% of data}
        \label{fig:convergence_half}
    \end{subfigure}
    
    \caption{
    \textbf{Training Efficiency Comparison in Full and Low-Data Regimes.}
    Success rate versus training steps for FlowVLA and the baseline.
    Our method converges dramatically faster and reaches a higher peak performance across both the full dataset \textbf{(a)} and a data-scarce setting \textbf{(b)}. The performance gap widens significantly with limited data, highlighting the superior sample efficiency of our approach.
    }
    \label{fig:convergence}
\end{figure}

To isolate and evaluate the impact of our Visual CoT pre-training, we conduct a direct comparison between FlowVLA and its foundational baseline, UniVLA. Figure~\ref{fig:convergence} illustrates FlowVLA's dramatic advantage in training and sample efficiency. In the full-data setting (Figure~\ref{fig:convergence}(a)), FlowVLA proves vastly more sample-efficient, reaching the baseline's peak performance (0.64) with only \textbf{one-third of the training steps} (2k vs. 6k) while also achieving a higher final success rate of 0.73.

This efficiency advantage is amplified in the more challenging low-data regime (Figure~\ref{fig:convergence}(b)). Here, the performance gap widens substantially. FlowVLA not only achieves a \textbf{55\% higher peak success rate} relative to the baseline (0.48 vs. 0.31) but also surpasses the baseline's peak performance in just 1k steps. This substantial improvement in sample efficiency validates our core hypothesis: by requiring the model to explicitly reason about motion via a visual chain-of-thought, FlowVLA benefits from a powerful inductive bias. This simplifies the learning of physical dynamics from raw pixels, leading to a more effective and robust learning process, particularly when data is limited.

\subsection{Ablation Studies (Q4)}
Finally, we conduct a series of ablation studies to understand the contribution of each key component in our framework. The results, summarized in Table~\ref{tab:ablations}, are evaluated on the LIBERO-10 benchmark.

We first remove the entire Visual Chain-of-Thought (CoT) structure, which causes our model to degenerate into the UniVLA baseline. As shown in Table~\ref{tab:ablations}, the success rate drops sharply from 73.0\% to 64.0\%. This significant 9-point drop confirms that the explicit, step-by-step reasoning process, where the model first thinks about ``how to move'' before predicting the outcome, is the primary driver of our model's performance gain.

Next, we investigate the importance of direct supervision for the intermediate reasoning step. In this variant, we retain the interleaved visual-flow sequence structure but remove the optical flow loss during training, meaning the model is not explicitly guided to generate physically correct flows. The performance degrades to 69.5\%. This result indicates that while the interleaved architecture provides a useful structural prior, direct supervision is crucial to prevent the model from generating arbitrary or collapsed representations for the intermediate step (\textbf{$f_t$}). The supervision ensures the ``thought'' is physically grounded.

Finally, we validate the core design of interleaving information. 
We restructure the input sequence into the format \textbf{$v_0, v_1, \dots, f_0, f_1, \dots$}, where all visual frames are grouped first, followed by all corresponding flow frames.
This configuration leads to a severe performance collapse, with the success rate plummeting to 49.4\%. This is because the model can no longer leverage the predicted motion \textbf{$f_t$} to inform the generation of the next state \textbf{$v_{t+1}$} in a causal, forward-looking manner. This result provides strong evidence that the ``interleaved, step-by-step causal chain'' (\textbf{$v_t \rightarrow f_t \rightarrow v_{t+1}$}) is essential for effective planning and action generation.

\begin{table}[h!]
\centering
\caption{
    Ablation studies on the LIBERO-10 benchmark. We evaluate the importance of our key design choices: the Visual CoT structure, the flow supervision loss, and the interleaved sequence format. The full FlowVLA model is shown for comparison.
}
\label{tab:ablations}
\begin{tabular}{lc}
\toprule
\textbf{Configuration} & \textbf{Success Rate (\%)} \\
\midrule
\textbf{FlowVLA (Ours, Full Model)} & \textbf{73.0} \\
\midrule
\textit{Ablations:} & \\
1. w/o CoT (degenerates to baseline) & 64.0 \\
2. w/o Flow Loss & 69.5 \\
3. Grouped Sequence & 49.4 \\
\bottomrule
\end{tabular}
\end{table}

\section{Related Work}
\label{sec:related_work}

\subsection{Vision-Language-Action (VLA) Models}
The dominant paradigm for creating generalist robot agents is the Vision-Language-Action (VLA) model~\cite{zitkovich2023rt, kim2024openvla, black2024pi0visionlanguageactionflowmodel, song2025acceleratingvisionlanguageactionmodelintegrated, song2025reconvla}. These models extend large, pre-trained Vision-Language Models (VLMs) by fine-tuning them on extensive robotics datasets~\cite{o2024open}. Architectures like RT-2~\cite{zitkovich2023rt}, CEED-VLA~\cite{song2025ceedvlaconsistencyvisionlanguageactionmodel} and OpenVLA~\cite{kim2024openvla} treat action generation as a sequence modeling problem, directly mapping visual and textual inputs to discretized action tokens. Other recent works have focused on improving the action representation itself, using techniques like diffusion policies~\cite{chi2023diffusion} or flow matching~\cite{black2024pi0visionlanguageactionflowmodel}. While this end-to-end approach has demonstrated remarkable generalization, it often treats the environment's physical dynamics as a ``black box''. The policy is learned reactively, without an explicit, underlying model of how the world functions or evolves. FlowVLA diverges from this standard VLA formulation by prioritizing world understanding over immediate action generation. Its pre-training objective is not to learn a policy ($v_t \rightarrow a_t$), but to build a robust world model by learning the physical transition function of the environment ($v_t \rightarrow v_{t+1}$). This ``dynamics-first'' approach establishes a solid foundation of physical knowledge before it is adapted for downstream control.

\subsection{World Models for Robotics}
The concept of a world model, which learns a model of its environment to plan or imagine future outcomes~\cite{ha2018world}, is increasingly vital in robotics. Recent works have leveraged this idea for policy learning. For example, some models use video prediction as a form of self-supervised pre-training to improve downstream task performance~\cite{wang2025unifiedvisionlanguageactionmodel, wu2023unleashing}. Others, like WorldVLA~\cite{cen2025worldvla}, propose architectures that jointly learn to predict both the next frame and the next action, creating a tight loop between prediction and control. A common thread in these approaches is the direct prediction of the next frame, modeling the transition as $v_t \rightarrow v_{t+1}$. However, this direct objective forces a single network to simultaneously handle two distinct problems: understanding static scene properties (appearance, texture, lighting) and modeling complex physical dynamics (motion, interaction, causality). This entanglement can result in inefficient learning and physically implausible predictions, such as blurry or distorted futures. In contrast, FlowVLA avoids this entanglement with its Visual Chain of Thought framework. We decompose the prediction into a ``frame $\rightarrow$ flow $\rightarrow$ frame'' reasoning process. By forcing the model to first predict an intermediate optical flow field ($f_t$), we explicitly decouple the learning of dynamics (\textit{how} things move) from appearance (\textit{what} they look like), resulting in a more causally-grounded world model.

\subsection{Embodied Reasoning for Robotics}
To move beyond simple reactive policies, a significant line of research has focused on endowing agents with explicit reasoning capabilities. These approaches can be broadly categorized. One category focuses on high-level semantic reasoning, where models generate linguistic or abstract plans. For instance, ECoT~\cite{zawalski2024robotic} and ThinkAct~\cite{huang2025thinkactvisionlanguageactionreasoningreinforced} leverage Chain-of-Thought prompting to generate textual sub-goals that guide the agent's behavior. A second category focuses on mid-level geometric reasoning, where models produce intermediate spatial representations to guide actions. MolmoAct~\cite{lee2025molmoactactionreasoningmodels}, for example, generates depth maps and 2D end-effector trajectory traces as part of its ``Action Reasoning'' pipeline to make planning more concrete. FlowVLA introduces a more fundamental form of reasoning: low-level physical reasoning. Unlike high-level semantic or geometric planning, our Visual CoT operates at the pixel level. By predicting the dense optical flow field, it learns a general, causal model of the world's dynamics, independent of any specific task or action. This provides a foundational understanding of physics that is complementary to, and arguably a prerequisite for, effective high-level control.

\section{Conclusion}

We proposed the Visual Chain of Thought (Visual CoT) as a new paradigm for world model learning, instantiated in FlowVLA. By decomposing prediction into an explicit \emph{motion–then–appearance} reasoning sequence $v_t \rightarrow f_t \rightarrow v_{t+1}$, our model learns physically grounded representations that align with the demands of downstream action prediction. 
We introduce a two-stage training paradigm. At the first stage, we pre-train the world model with Visual CoT to build motion-coherent and physically plausible dynamics knowledge from video data. At the second stage, we fine-tune the policy to adapt this knowledge to generate precise robot actions. This design directly addresses the gap between pre-training and fine-tuning in VLA models, leading to improved sample efficiency and robust performance.  
Experiments on simulation and real-world manipulation benchmarks validate the effectiveness of this motion-first approach, confirming that explicit motion reasoning is a powerful inductive bias for bridging perception and control in generalist robotics.

\clearpage
\bibliography{main}

\end{document}